\documentclass[sn-mathphys]{sn-jnl}

\usepackage{epstopdf}
\usepackage{subfigure}
\usepackage{float}
\usepackage{color}

\jyear{2021}%

\theoremstyle{thmstyleone}%
\theoremstyle{thmstyletwo}%

\theoremstyle{thmstylethree}%

\begin{document}

\title[Multiresolution Feature Guidance Based Transformer for Anomaly Detection]{Multiresolution Feature Guidance Based Transformer for Anomaly Detection}


\author[1]{\fnm{Shuting} \sur{Yan}}\email{yanst61@gmail.com}

\author*[1]{\fnm{Pingping} \sur{Chen}}\email{ppchen.xm@gmail.com}

\author[1]{\fnm{Honghui} \sur{Chen}}\email{chh5840996@gmail.com}

\author[1]{\fnm{Huan} \sur{Mao}}\email{maohuan980202@gmail.com}

\author[1]{\fnm{Feng} \sur{Chen}}\email{chenf@fzu.edu.cn}

\author[1]{\fnm{Zhijian} \sur{Lin}}\email{zlin@fzu.edu.cn}

\affil*[1]{\orgdiv{Department of Physics and Information Engineering}, \orgname{Fuzhou University}, \orgaddress{\city{Fuzhou}, \postcode{350108}, \country{China}}}




\abstract{Anomaly detection is represented as an unsupervised learning to identify deviated images from normal images. In general, there are two main challenges of anomaly detection tasks, i.e., the class imbalance and the unexpectedness of anomalies. In this paper, we propose a multiresolution feature guidance method based on Transformer named \emph{GTrans} for unsupervised anomaly detection and localization. In \emph{GTrans}, an Anomaly Guided Network (AGN) pre-trained on ImageNet is developed to provide  surrogate labels for features and tokens. Under the tacit knowledge guidance of the AGN, the anomaly detection network named \emph{Trans} utilizes Transformer to effectively establish a relationship between features with multiresolution, enhancing the ability of the \emph{Trans} in fitting the normal data manifold. Due to the strong generalization ability of AGN, \emph{GTrans} locates anomalies by comparing the differences in spatial distance and direction of multi-scale features extracted from the AGN and the \emph{Trans}. Our experiments demonstrate that the proposed \textit{GTrans} achieves state-of-the-art performance in both detection and localization on the MVTec AD dataset. \textit{GTrans} achieves image-level and pixel-level anomaly detection AUROC scores of 99.0\% and 97.9\% on the MVTec AD dataset, respectively.}




\maketitle

\section{Introduction}\label{sec1}

Anomaly detection aims to identify samples that do not match the expected pattern or dataset. Anomaly detection techniques have been extensively studied in a variety of research and application domains, including industrial inspection\cite{liu2021defect,wu2021automatic,he2021unsupervised} and medical applications\cite{zhou2020encoding}. With the continuous development of the industrial field, customers need to place various sensors for continuous monitoring of equipment condition and detect anomalies. The traditional manual visual detection method has recently become unsatisfactory because it is susceptible to the influence of manual experience and subjective factors. Therefore, anomaly detection has gradually become valuable in computer vision, attracting high attention in different fields \cite{zhao2021vehicle,zhou2021immune,zheng2021deep}. In general, class imbalance problem\cite{qian2020dr} and the unexpectedness of anomalies are the two most common challenges of anomaly detection tasks. Anomalies are extremely rare in the industrial scene, which implies the number of aberrant samples obtained is quite low, resulting in a serious class imbalance problem. In addition, anomalies are always unexpected. It is hard to predict their location and size or even determine if anomalies occur at all. As a result, modeling all anomalies or even predicting all abnormalities that never occur is impractical on a few samples.

	\begin{figure}[!t]
		\centering
			\includegraphics[width=10cm]{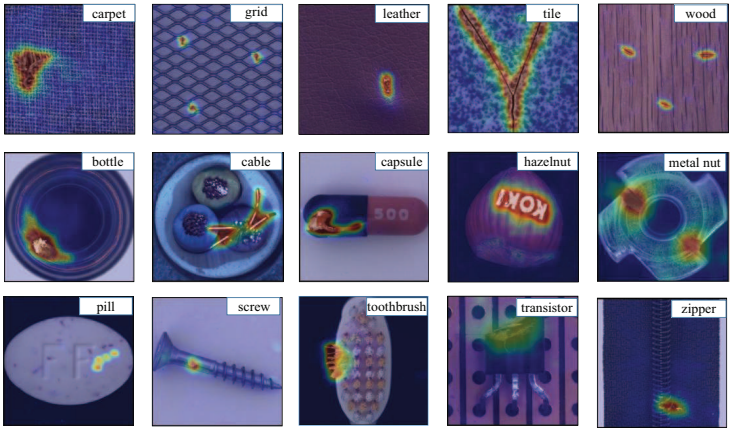}
		\caption{Visual results from the MVTec AD datasets. Superimposed on the images are the anomaly localization map from \textit{GTrans}. Red areas correspond to
		the located anomalies, whereas the blue areas indicate the normality regions.}
		\label{FIG:1}
	\end{figure}

\begin{figure*}[!t]
		\centering
		\includegraphics[width=12cm]{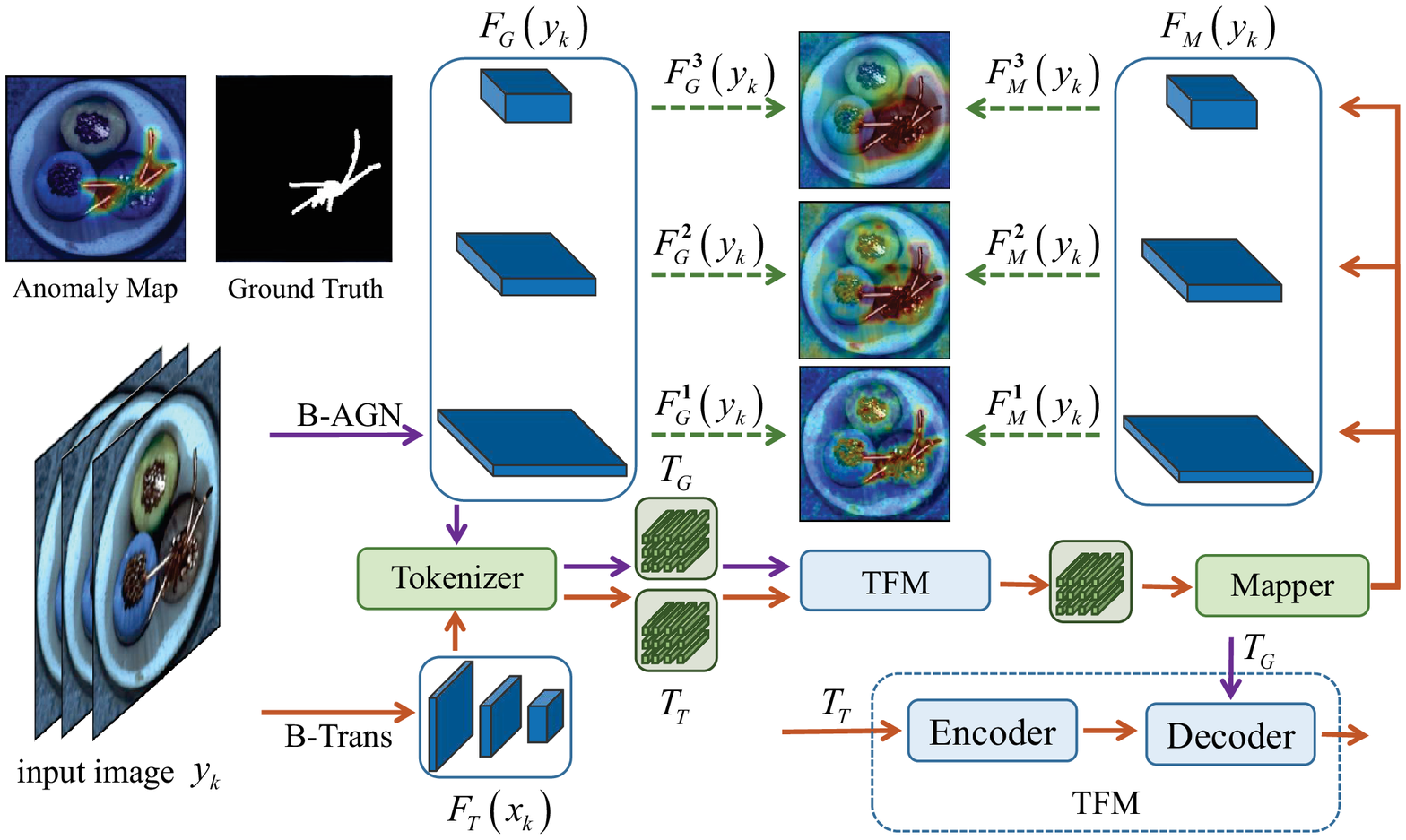}
		\caption{The overall architecture of the \emph{GTrans}. The solid purple arrow represents the processing of the AGN, and the solid orange arrow represents the processing of the \emph{Trans} network. Given an input image \textit{$y_{k}$}, we extract groups of multiresolution features \textit{${F_G}\left( {{y_k}} \right)$} and \textit{${F_T}\left( {{y_k}} \right)$} from B-AGN and B-Trans, where B-AGN and B-Trans represent  the backbone of AGN and \emph{Trans}, respectively. \textit{${F_G}\left( {{y_k}} \right)$} and \textit{${F_T}\left( {{y_k}} \right)$} are input into Tokenizer to obtain groups of tokens \textit{${T_G}$} and \textit{${T_T}$}. AGN guide the decoder module of TFM to enhance the fitting ability of \emph{Trans} in normal data manifold. The mapper module maps the token output by TFM into original critical layers. \emph{GTrans} locates anomalies by comparing the differences in spatial distance and direction of multi-scale features extracted from the AGN and the \emph{Trans}. Dotted green arrows represent the process of generating an anomaly map.}
		\label{FIG:2}
	\end{figure*}

The existing anomaly detection tasks\cite{perera2019ocgan,gong2019memorizing} focused on the classification of single or multiple categories at image-level. The current category to be detected is normal, and non-category are abnormal. However, in contrast to classification tasks, it is hard to train a model with full supervision for anomaly detection due to the lack of a large number of abnormal samples. \cite{an2015variational,ruff2018deep} trained models in normal category, and samples were judged as anomalous when they show a large difference from the trained normal samples in the test stage.

	Recently,  the anomaly detection tasks had confronted new challenges. To better monitor and process the anomalies, it is required not only to pick out the anomalous images but also to locate the anomalous regions. Bergmann et al.\cite{bergmann2019mvtec} proposed the MVTec AD dataset to provide benchmarks for anomaly detection and localization. The pixel-level methods\cite{pidhorskyi2018generative, schlegl2017unsupervised, perera2019ocgan, akcay2018ganomaly, schlegl2019f} exploited deep convolutional autoencoder and generative model such as Generative Adversarial Network (GAN) and Variational Autoencoder (VAE), respectively. \cite{gong2019memorizing, fei2020attribute, abati2019latent} attempted to learn the feature representation of the normal samples from scratch and the model trained under the normal data manifold is hard to  reconstruct the abnormal image. It may result  in a large per-pixel reconstruction error and then a higher anomaly score. However, these methods showed a tremendous potential for misdetection due to  low resolution of the reconstructed images and the strong generalization ability of the model.
	
	Fortunately, it was found that using a pre-trained network could be a potential mode to learn feature representation for small sample datasets. Cohen et al.\cite{cohen2020sub} utilized a group of features extracted from a deep pre-trained network on anomaly-free images to train their model. \cite{cohen2020sub} improved performance gain, but their model relied on many sub-images of training data, resulting in expensive computations. To evade these limitations, Bergmann et al.\cite{bergmann2020uninformed} proposed a student-teacher network to implicitly model the distribution of features extracted from normal images. The idea of the student-teacher network was that the student network had poor generalization ability to abnormal data manifolds and then made wrong judgments. For knowledge transfer, \cite{bergmann2020uninformed} only imitated the last layer in the teacher network for knowledge distillation without making full use of the information of intermediate layers. To fully exploit  the intermediate features of the teacher network, \cite{salehi2021multiresolution} proposed a novel knowledge distillation method that distilled the comprehensive knowledge of the pre-trained network at several critical layers to the trained network. It can provide the significance of multi-scale information in computer vision algorithms. Although features were extracted in several critical layers in \cite{salehi2021multiresolution}, the information interaction between multi-scale features is fragile.
		
	Lately, Transformer\cite{vaswani2017attention} had achieved great success in image classification and recognition, which proved the feasibility of Transformer in the information interaction of features with multiresolution. Inspired by this, we propose a multiresolution feature guidance method based on Transformer named \emph{GTrans} for unsupervised anomaly detection and localization.
In our method, the Anomaly Detected Network based on Transformer named \emph{Trans} utilizes Transformer to effectively establish relationship between features with multiresolution. Under the tacit knowledge transfer of the Anomaly Guided Network (AGN), \emph{Trans} enhances the fitting ability on the normal data manifold.
Our intuition is that when abnormal images are input into the \emph{Trans}, \emph{Trans} cannot judge such accidents, resulting in large abnormalities in the whole image. Due to the strong ability of \emph{Trans} in obtaining interaction information of features, the abnormal regions will receive great attention, conducive to the realization of pixel-level localization of the anomalies.
	Furthermore, we define a novel function to compute the anomaly score according to the difference of features extracted by AGN and \emph{Trans} in the spatial distance and direction. The function can effectively capture the information of the combination of multi-scale features to obtain an accurate heat map of anomalous regions. We evaluate our method on MVTec AD dataset and then achieve superior performance in both localization and detection. Figure \ref{FIG:1} shows visualized results of our method on the MVTec AD dataset. Our main contributions are summarized as follows:

    \begin{enumerate}
		\item  We propose a multiresolution feature guided method based on Transformer, referred as \emph{GTrans} for unsupervised\footnote{Since training set in  \textit{GTrans} only contains  normal images without any labels, such  data setup can be generally considered as unsupervised \cite{gudovskiy2022cflow}.} anomaly detection and localization. \emph{GTrans} can improve the ability in information interaction between features with multiresolution.

        \item  We develop an anomaly guided network (AGN) to provide surrogate labels of features with multiresolution. Our model can be trained on normal images entirely without additional data augmentation.
		
		\item  We define a novel function to generate an anomaly map by exploiting the anomaly maps with multiresolution in spatial distance and direction.

        \item  Experimental results show that our proposed \emph{GTrans} achieves state-of-the-art performance on MVTec AD dataset, which demonstrates the significance of  information interaction for anomaly detection and localization.
	
	\end{enumerate}

\section{Related Work}\label{sec2}

    Anomaly detection has attracted  a lot of attention in the last decades. We provide an extensive overview of anomaly detection techniques next. The research related to our work can categorize into two classes. First, we present an overview of anomaly detection and segmentation methods, which can be  categorized  into reconstruction-based methods and embedding similarity-based methods. Second, we briefly explain Transformer architectures and show the application of Transformer in anomaly detection.

\subsection{Anomaly Detection and Segmentation}\label{subsec21}

\subsubsection{Reconstruction-based methods}\label{subsubsec211}

Reconstruction-based methods like autoencoders (AEs)\cite{bergmann2018improving, gong2019memorizing, fei2020attribute, abati2019latent}, variational autoencoders (VAEs)\cite{sato2019predictable, liu2020towards, lu2018anomaly, an2015variational} or generative adversarial networks (GANs)\cite{sabokrou2018adversarially, pidhorskyi2018generative, schlegl2017unsupervised, perera2019ocgan, akcay2018ganomaly, schlegl2019f} attempted to model the sample without abnormality and defect from scratch. 
	
	The idea of AE-based methods was that if the model were trained on a dataset containing only normal samples, the reconstructed image would approach the normal sample regardless of whether the input images are normal samples or abnormal samples. By comparing the pixel error of the original image and the reconstructed image, the model could judge whether the input image is abnormal or not and even locate the abnormal regions. To learn potential features better, SSIM-AE\cite{bergmann2018improving} used SSIM as loss function and anomaly measure to compare input and reconstructed images. ARNet\cite{fei2020attribute} learned the semantic feature embeddings related to the erased attributes by forcing the network to restore the original image. MemAE\cite{gong2019memorizing} proposed an autoencoder with a memory module to explicitly suppress the generalization capability of the autoencoder.
	
	VAE-based methods were also found to be used in anomaly detection and localization. In VAE-based methods, the probabilistic encoder and decoder both parameterized an isotropic normal distribution in the latent variable space and the original input variable space, respectively. The model endeavored to find the probability distribution conforming to the normal sample and used the reconstructed probability as the anomaly scores to generate an attention map in the potential space to detect the anomalies. However, VAE-based methods were not automatically superior to traditional autoencoder methods in general.
	
	GAN-based approaches utilized the discriminator to detect the slight disturbance in the reconstruction of abnormal images, prompting the generator to extract sample information from the potential space to reconstruct the image. AnoGAN\cite{schlegl2017unsupervised} proposed a novel anomaly scoring scheme based on the mapping from image space to a latent space. OCGAN\cite{perera2019ocgan} utilized a dual latent space learning process to constrain the latent space of generators and discriminators to represent the specific categories exclusively. GANomaly\cite{akcay2018ganomaly} and f-AnoGan\cite{schlegl2019f} added an additional encoder to the generator to reduce the inference time of AnoGan\cite{schlegl2017unsupervised}.
	
	Methods based on reconstruction were intuitive and explicative. However, these methods either failed to detect anomalies due to the strong generalization ability of deep models\cite{bergmann2018improving, gong2019memorizing} or failed in one-class settings\cite{schlegl2017unsupervised, perera2019ocgan, akcay2018ganomaly, schlegl2019f}.

\subsubsection{Embedding similarity-based methods}\label{subsubsec212}

Embedding similarity-based methods mainly mapped the meaningful vector of the sample extracted from the pre-trained network to the high-dimensional feature space and judged the possibility of anomalies by computing the distance between the test sample and the normal sample in the feature space. Feature space expressed a higher level and more abstract information than image space. Embedding similarity-based methods could be categorized as cluster-based, embedding patch-based and knowledge-based methods.
	
	Cluster-based methods used K-Nearest Neighbor (KNN)\cite{eskin2002geometric}, K-means\cite{hartigan1979algorithm} or Principal Component Analysis (PCA)\cite{teh2021expect} to match the best approximate features from a memory-bank of nominal features or similar feature sets extracted from a pre-trained network to locate anomalies. However, the inference speed in the test phase of cluster-based methods was linearly related to the size of feature sets, which was limited in practical application.
	
	Embedding patch-based methods tried to repair abnormal images by patch extracted from normal images. Patch SVDD\cite{yi2020patch} and CutPaste\cite{li2021cutpaste} applied self-supervised learning to anomaly detection. Unlike Deep SVDD\cite{ruff2018deep}, Patch SVDD\cite{yi2020patch} inspected the image at the patch level, and each patch corresponds to a point in the feature space. CutPaste\cite{li2021cutpaste} proposed a data augmentation strategy that cut an image patch and pasted it at a random location of an image. PaDiM\cite{defard2021padim} described each patch location with a Gaussian distribution and modeled the correlation between semantic layers. However, self-supervised representation typically underperformed those learned from large supervised datasets such as ImageNet.
	
	Knowledge-based methods mainly utilized the difference in generalization ability between teacher and student networks in a teacher-student framework to locate anomalies. The teacher network had a good performance and strong generalization ability, while the student network showed strong representation ability only in a single class by imitating the behavior of the teacher network. Bergmann et al.\cite{bergmann2020uninformed} were the first to introduce a teacher-student framework in the field of unsupervised anomaly detection and localization. Anomaly scores of \cite{bergmann2020uninformed} were derived from the predictive variance and regression error of an ensemble of student networks. On this basis, \cite{salehi2021multiresolution} and \cite{wang2021student} extended the multi-scale critical layers of knowledge extraction on the VGG and ResNet networks respectively, and achieved good performance. However, such methods either only extracted the single-layer information without making full use of the information in intermediate layers\cite{bergmann2020uninformed} or treated critical layers with multiresolutions equally while ignoring the importance of information interaction of multi-scale critical layers\cite{salehi2021multiresolution}\cite{wang2021student}.

\subsection{Transformer in Vision}\label{subsubsec22}

Ref. \cite{vaswani2017attention} first proposed Transformer, a model relying entirely on an attention mechanism to describe global dependencies between input and output, in the field of Natural Language Processing (NLP). Transformer did not use Recurrent Neural Network (RNN) such as Long Short-Term Memory (LSTM)\cite{hochreiter1997long} and instead utilized a stack of multi-headed attention blocks to accomplish NLP tasks excellently. The specific structure of the Transformer can be found in the original article\cite{vaswani2017attention}.
	
	Transformer had achieved the most advanced performance in many NLP tasks and had become the preferred model for NLP tasks. Computer vision and NLP are merging as more efficient structures emerge. Due to the computational efficiency and scalability of Transformer, Transformer had been explored in computer vision and had become a new research direction.
	
	Recently, Transformer had achieved good performance in image classification\cite{dosovitskiy2020image}, object detection\cite{carion2020end}\cite{zhu2020deformable} and image segmentation\cite{chen2021transunet}. In general, there were two main model architectures for the adoption of Transformer in computer vision. One is a pure Transformer structure, and the other is a hybrid structure combining convolution neural network (CNN) and Transformer. Moreover, Vision Transformer\cite{dosovitskiy2020image} built a pure Transformer structure, dividing the input image into square uniform patches with \textit{$16 \times 16$} patch size. For each patch, the linear transformation was performed for dimensionality reduction and location information was embedded. Then, the projection information and location information was input into the Transformer to achieve image classification. \cite{chen2021transunet} proposed TransUNet, which merited both Transformers and U-Net\cite{ronneberger2015u}, to achieve medical image segmentation. TransUNet\cite{chen2021transunet} utilized Transformer for encoding tokenized image patches from a feature map extracted by CNN, and the decoder upsampled the encoded features combined with the high-resolution CNN feature maps to enable precise localization.

	For anomaly detection and localization, InTra\cite{pirnay2021inpainting} used a deep Transformer network consisting of a simple stack of multi-headed self-attention blocks to detect anomalies. Experiments showed that pure Transformer architecture is more efficient and scalable than traditional CNN in both model size and computational scale, while hybrid architecture performed better than pure Transformer at the smaller model size. Therefore, our method employ  the hybrid structure combining CNN and Transformer to detect and locate anomalies and achieved excellent results.

\section{Methodology}\label{sec3}

We propose a multiresolution feature guidance method based on Transformer named \emph{GTrans} for anomaly detection and localization. \emph{GTrans} consists of four components: 1) AGN: Anomaly guided network. 2) \emph{Trans}: Anomaly detected network based on Transformer. 3) Learning of the normality. 4) Computation of the anomaly map. The overall architecture of \emph{GTrans} is shown in Figure \ref{FIG:2}. Given a testing dataset \textit{${D_{test}}{=}\left\{ {{y_1},{y_2},...,{y_M}} \right\}$} consisting both anomaly and normal images. Given an input image \textit{$y_{k}\in {\mathbb{R}^{W\times{H}\times{C}}}$}, \textit{$k \in \left[ {1,M} \right]$} of width \textit{H}, height \textit{W}, and the number of channels \textit{C}, we extract groups of features with multiresolution \textit{${F_G}\left( {{y_k}} \right) = \left\{ {F_G^1\left( {{y_k}} \right), F_G^2\left( {{y_k}} \right),...,F_G^L\left( {{y_k}} \right)} \right\}$} and \textit{${F_T}\left( {{y_k}} \right) = \left\{ {F_T^1\left( {{y_k}} \right), F_T^2\left( {{y_k}} \right),...,F_T^L\left( {{y_k}} \right)} \right\}$} from AGN pre-trained on ImageNet and the backbone of \emph{Trans}, respectively, where \textit{L} represents total number of critical layers. \textit{${F_G}\left( {{y_k}} \right)$} and \textit{${F_T}\left( {{y_k}} \right)$} are input into Tokenizer to obtain groups of tokens \textit{${T_G}{\rm{ = }}\left\{ {T_G^1, T_G^2,...,T_G^L} \right\}$} and \textit{${T_T}{\rm{ = }}\left\{ {T_T^1, T_T^2,...,T_T^L} \right\}$}, respectively. \emph{Trans} utilizes Transformer to establish a relationship between features with multiresolution. AGN guides the decoder module of TFM to enhance the fitting ability of \emph{Trans} in a normal data manifold. The mapper module maps the token output by TFM into original critical layers. In training, the parameters of AGN are frozen and the critical layers extracted by AGN as regression targets of the \emph{Trans}. In testing, \emph{GTrans} locates anomalies by comparing the differences in spatial distance and direction of multi-scale features extracted from the AGN and the \emph{Trans}.

In the following, we describe the four components of \emph{GTrans} in detail.

\subsection{Anomaly Guided Network}\label{subsec31}

We propose a novel network called Anomaly Guided Network (AGN) for unsupervised anomaly detection and localization. AGN is a convolution network pre-trained on ImageNet (e.g., a ResNet-50-2 pre-trained on ImageNet), it can exhibit strong ability in feature representation when the sample quantity is small as ever. The guiding of AGN comes from two aspects. First, the token extracted from AGN as surrogate labels to be learned of \emph{Trans}. Second, the feature maps extracted by AGN as regression targets of the \emph{Trans} during the learning of the normality, enhancing the fitting ability of \emph{Trans} in normal data manifolds.

\subsubsection{Feature Extraction}\label{subsubsec311}

Given a training dataset \textit{${D_{train}}{=}\left\{ {{x_1},{x_2},...,{x_N}} \right\}$} consisting only of normal images. Given an input image \textit{$x_{k}\in {\mathbb{R}^{W\times{H}\times{C}}}$}, \textit{$k \in \left[ {1,N} \right]$} of width \textit{H}, height \textit{W}, and the number of channels \textit{C}. Our goal is to create an ensemble of feature group \textit{${F_G}\left( {{x_k}} \right) = \left\{ {F_G^1\left( {{x_k}} \right),F_G^2\left( {{x_k}} \right),...,F_G^L\left( {{x_k}} \right)} \right\}$} to detect anomalies of various sizes, the \textit{l}-th critical layer of AGN extracts a feature map \textit{$F_G^l({x_k}) \in {\mathbb{R}^{{w_l}{\rm{ \times }}{h_l}{\rm{ \times }}{c_l}}}$} for \textit{$l \in \left[ {1,L} \right]$} , where \textit{$h_{l}$} and \textit{$w_{l}$} represent the height and width of the feature map, \textit{$c_{l}$} represents the number of channels.

\subsubsection{Token Extraction}\label{subsubsec312}

    Inspired by \cite{zhang2019latentgnn,liang2018symbolic,chen2019graph}, we utilize a module designated Tokenizer, containing point-wise convolutions and spatial attention, to extract visual tokens because a few visual tokens are sufficient to represent the entire image.

 We reshape the input feature map \textit{$F_G^l({x_k}) \in {\mathbb{R}^{{w_l}{\rm{ \times }}{h_l}{\rm{ \times }}{c_l}}}$} into a sequence of flattened two-dimensional feature map \textit{$F_{GK}^l({x_k}) \in {\mathbb{R}^{\left( {{w_l} \cdot {h_l}} \right) \times {c_l}}}$} before entering the Tokenizer. Tokenizer uses two point-wise convolutions \textit{$L_{g}$} and \textit{$L_{d}$} to process each pixel of two-dimensional feature map \textit{$F_{GK}^l({x_k})$} as follows:
	 \begin{equation}
	 	{G_l} = \frac{{{L_g}\left( {F_{GK}^l({x_k})} \right)}}{{\sqrt {{c_l}} }} \in {\mathbb{R}^{\left( {{w_l} \cdot {h_l}} \right) \times g}}
	 \end{equation}
and
 	\begin{equation}
 		{V_l} = {L_d}\left( {F_{GK}^l({x_k})} \right) \in {\mathbb{R}^{\left( {{w_l} \cdot {h_l}} \right) \times d}}.
 	\end{equation}

 	Here, \textit{$L_{g}$} maps each pixel of feature map in \emph{l}-th critical layer to \textit{g} semantic groups \textit{$G_{l}$}, while \textit{$L_{d}$} realizes the information interaction between channels of the feature map by raising or reducing dimension. \textit{$V_{l}$} represents the high-dimensional feature representation of \emph{l}-th critical layers. For each semantic group, spatial attention is computed for the sequences \textit{$G_{l}$} as a weighted average over \textit{$V_{l}$} as follow:
	\begin{equation}
		{T^l} = {\rm{softmax}}{({G_l})^T}{V_l} \in {R^{d \times g}},
	\end{equation}
where \textit{$\rm{softmax}( \cdot )$} obtains the spatial attention weight of each semantic group.

In general, we input \textit{${F_G}\left( {{x_k}} \right)$} into the Tokenizer to obtain a group of visual tokens \textit{${T_G}{\rm{ = }}\left\{ {T_G^1,T_G^2,...,T_G^L} \right\}$} with different resolutions. For anomaly detection, {${T_G}$} and \textit{${F_G}\left( {{x_k}} \right)$} provide guidance in decoder modules and the learning of the normality, respectively.

\subsection{\emph{Trans}}\label{subsec32}

We propose a anomaly detected network based on Transformer named \emph{Trans}. \emph{Trans} improves the ability in information interaction between features with multiresolution. In the following, we describe the steps of \emph{Trans} in detail.

\subsubsection{Extraction of feature and token}\label{subsubsec321}

To effectively obtain the tacit knowledge of AGN, we rely on architecture that the dimension of the features of \emph{Trans} is aligned with that of AGN. \emph{Trans} and AGN have the same backbone, but \emph{Trans} has not been pre-trained. In CNN, top layers produce low-resolution, semantically strong features, while bottom layers produce high-resolution, semantically weak ones. Given an input image \textit{$x_{k}\in {\mathbb{R}^{W\times{H}\times{C}}}$} of width \textit{H}, height \textit{W}, and the number of channels \textit{C}. To detect and locate anomalies with various sizes, we extract a group of features \textit{${F_T}\left( {{x_k}} \right) = \left\{ {F_T^1\left( {{x_k}} \right),F_T^2\left( {{x_k}} \right),...,F_T^L\left( {{x_k}} \right)} \right\}$} at various layers. We input \textit{${F_T}\left( {{x_k}} \right)$} into the Tokenizer to obtain a group of visual tokens \textit{${T_T}{\rm{ = }}\left\{ {T_T^1,T_T^2,...,T_T^L} \right\}$} with multiresolutions.

\subsubsection{TFM}\label{subsubsec322}

This module aims to establish information interaction of spatially distant concepts for visual tokens of different semantic groups at different levels. We employ the encoder-decoder structure of standard Transformer with the following changes: (1) Omitting position embedding. \cite{islam2020much} provided convincing evidence that CNNs do indeed rely on and learn information about spatial positioning. Therefore, the convolution layer computes learnable weights of query \textit{${q_e} \in {\mathbb{R}^{d \times \left( {g \cdot l} \right)}}$}, key \textit{${k_e} \in {\mathbb{R}^{d \times \left( {g \cdot l} \right)}}$}, and value \textit{${v_e} \in {\mathbb{R}^{d \times \left( {g \cdot l} \right)}}$} without additional position embedding. (2) Using a non-linearity activation function and two point-wise convolutions replace position-wise feed-forward networks of the standard Transformer. (3) Extracting a group of visual tokens \textit{${T_G}{\rm{ = }}\left\{ {T_G^1,,T_G^2,...,T_G^L} \right\}$} from AGN as the input of the decoder in Transformer.


%

	\begin{figure}[!t]
		\centering
			\includegraphics[width=12cm]{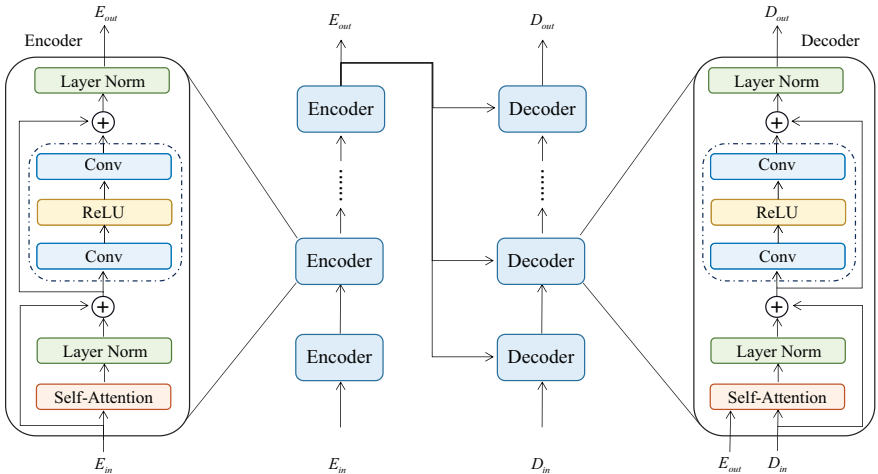}
		\caption{Overview of the TFM architecture.}
		\label{FIG:31}
	\end{figure}


    We name the module modified by standard Transformer as TFM. Figure \ref{FIG:31} illustrates the architecture of the TFM. As can be seen, TFM is composed of a stack of \emph{S} encoders and decoders. Each encoder and decoder has three sub-layers, respectively.

The first is a self-attention mechanism, the second is a combination of a non-linearity activation function and two point-wise convolutions, and the third is a layer normalization.  We take an individual TFM block of \textit{$S = 1$} as an example below. We concatenate the group of visual tokens \textit{${T_G}$} and \textit{${T_T}$} into \textit{${E_{in}} \in {R^{d \times \left( {g \cdot l} \right)}}$} and \textit{${D_{in}} \in {\mathbb{R}^{d \times \left( {g \cdot l} \right)}}$}, respectively. We take \textit{${E_{in}}$} and \textit{${D_{in}}$} as the input of the encoder and decoder in TFM.

  In the encoder, we compute \textit{${q_e}$}, \textit{${k_e}$}, \textit{${v_e}$} via
	\begin{equation}
		{q_e} = {W_q}{E_{in}},{k_e} = {W_k}{E_{in}},{v_e} = {W_v}{E_{in}} \in {\mathbb{R}^{d \times \left( {g \cdot l} \right)}}
	\end{equation}
with learnable weight matrices \textit{${W_q},{W_k},{W_v} \in {\mathbb{R}^{d \times d}}$}. Then, self-attention is computed by a compatibility function of the \textit{${q_e}$} with corresponding \textit{${k_e}$} as a weighted average of the \textit{${v_e}$} as follows:
	\begin{equation}
		{A_e} = {v_e}{\rm{softmax}}\left( {\frac{{k_e^T{q_e}}}{{\sqrt d }}} \right) \in {\mathbb{R}^{d \times \left( {g \cdot l} \right)}},
	\end{equation}
where \textit{${A_e}$} is the output of self-attention mechanism in encoder.

	We employ a residual connection around the self-attention mechanism and layer normalization to get the output of layer normalization \textit{${E_a}$}:
	\begin{equation}
		{E_a} = {E_{in}} + LayerNorm\left( {{A_e}} \right) \in {\mathbb{R}^{d \times \left( {g \cdot l} \right)}},
	\end{equation}
where \textit{$LayerNorm\left(  \cdot  \right)$} represents the layer normalization.

    Then, we use a residual connection around the convolution layer, followed by layer normalization to get the output of encoder \textit{${E_{out}}$}:
	\begin{equation}
		{E_{out}} = LayerNorm\left( {{E_a} + {L_1}\sigma \left( {{L_2}{E_a}} \right)} \right) \in {\mathbb{R}^{d \times \left( {g \cdot l} \right)}},
	\end{equation}
where \textit{${L_1},{L_2} \in {R^{d \times d}}$} are point-wise convolutions, \textit{$\sigma \left(  \cdot  \right)$} is the \textit{$ReLU$} function.
	
	For the decoder, most methods\cite{dosovitskiy2020image,chen2021transunet} previously applied to computer vision tasks use only the encoder of the Transformer. However, if the self-attention mechanism is only established in the visual token extracted from Tokenizer in \emph{Trans}, the final model will easily fall into the problem of local optimal. Therefore, we employ an encoder-decoder structure of the Transformer and utilize visual tokens extracted from AGN as surrogate labels for the guidance of \textit{${E_{out}}$} to enhance the stability of the \textit{Trans}. We set
	\begin{equation}
		{q_d} = {W_q}{D_{in}},{k_d} = {W_k}{E_{out}},{v_d} = {W_v}{E_{out}} \in {\mathbb{R}^{d \times \left( {g \cdot l} \right)}},
	\end{equation}
	\begin{equation}
		{A_d} = {v_d}{\rm{softmax}}\left( {\frac{{k_d^T{q_d}}}{{\sqrt d }}} \right) \in {\mathbb{R}^{d \times \left( {g \cdot l} \right)}}.
	\end{equation}
	Here, \textit{${q_d}$} is the query computed from \textit{${D_{in}}$}. \textit{${k_d}$}, \textit{${v_d}$} are the key and value computed from the output of the encoder \textit{${E_{out}}$}. \textit{${A_d}$} is the output of self-attention mechanism in decoder.

Similarly to the encoder, first, we employ a residual connection around the self-attention mechanism and layer normalization to get the output of layer normalization \textit{${D_a}$} in the decoder. Second, we employ another residual connection around the convolution layer, followed by layer normalization to get the output of decoder \textit{$D_{out}$}. The computational process is as follows:
	\begin{equation}
		{D_a} = {D_{in}} + LayerNorm\left( {{A_d}} \right) \in {\mathbb{R}^{d \times \left( {g \cdot l} \right)}}
	\end{equation}
and
	\begin{equation}
		{D_{out}} = LayerNorm\left( {{D_a} + {L_1}\sigma \left( {{L_2}{D_a}} \right)} \right) \in {\mathbb{R}^{d \times \left( {g \cdot l} \right)}}.
	\end{equation}

\subsubsection{Mapping of feature and token}\label{subsubsec323}

To achieve pixel-level alignment between the outputs of AGN and the \textit{Trans}, we divide \textit{${D_{out}}$} into \textit{g} groups and map each group into the latent space of the same dimension as the \textit{${F_G}\left( {{x_k}} \right) = \left\{ {F_G^1\left( {{x_k}} \right),F_G^2\left( {{x_k}} \right),...,F_G^L\left( {{x_k}} \right)} \right\}$}, respectively. For the \textit{l}-th critical layer, we set
	\begin{equation}
		F_{M}^l({x_k}) = F_T^l({x_k}) + T_v^l{\rm{softmax}}{\left( {\frac{{{{\left( {X_q^l} \right)}^T}T_k^l}}{{\sqrt {{c_l}} }}} \right)^T},
	\end{equation}
where \textit{$F_{M}^l({x_k})$} is the final output of the \textit{l}-th critical layer, \textit{$X_q^l$} is the query computed from \textit{$F_G^l({x_k})$}. \textit{$T_k^l$}, \textit{$T_v^l$} are the key and value computed from the output of the encoder in TFM \textit{${E_{out}}$}.
	
    In general, the final output of \emph{Trans} is a group of critical layers \textit{${F_{M}}({x_k}){\rm{ = }}\left\{ {F_{M}^1({x_k}),F_{M}^2({x_k}),...,F_{M}^L({x_k})} \right\}$}.

\subsection{Learning of the normality}\label{subsec33}

In the following, we train the \textit{Trans} so that its output is as similar as possible to the AGN, and the parameters of AGN are frozen during the entire training phase. Given an input image  \textit{${x_k} \in {D_{train}}$}, we obtain a group of feature maps \textit{${F_G}({x_k}){\rm{ = }}\left\{ {F_G^1({x_k}),F_G^2({x_k}),...,F_G^L({x_k})} \right\}$} and \textit{${F_{M}}({x_k}){\rm{ = }}\left\{ {F_{M}^1({x_k}),F_{M}^2({x_k}),...,F_{M}^L({x_k})} \right\}$} extracted from AGN and \textit{Trans}, respectively. As indicated in \cite{yim2017gift}, \textit{Trans} can learn the distilled knowledge from AGN that is trained at a different task and guided by intermediate-level hints from AGN. \textit{Trans} can learn the intermediate-level hints from different semantic layers of AGN. It then takes the intermediate-level hints as targets of optimization procedure to intensify the complete knowledge transfer from AGN to \textit{Trans}. We define pixel-wise \textit{${L_2}$} loss function \textit{$p_{\left( {i,j} \right)}^l$} at position \textit{$\left( {i,j} \right)$} for $l \in \left[ {1,L} \right]$, $i \in \left[ {1,{w_l}} \right]$, $j \in \left[ {1,{h_l}} \right]$ as follow:
	\begin{equation}
		p_{\left( {i,j} \right)}^l\left( {{x_k}} \right) = \frac{1}{2}\left\| {F_G^l{{\left( {{x_k}} \right)}_{(i,j)}} - F_{M}^l{{\left( {{x_k}} \right)}_{\left( {i,j} \right)}}} \right\|_2^2.
	\end{equation}

	Here, \textit{$F_G^l\left( {{x_k}} \right)$} and \textit{$F_{M}^l\left( {{x_k}} \right)$} represent the feature map extracted from \emph{l}-th critical layers of AGN and \textit{Trans}. We get the total loss \textit{$L\left( {{x_k}} \right)$} of input image \textit{${x_k}$} by the weighted average of all pixels in each feature map of critical layers in sets \textit{$F_G^l\left( {{x_k}} \right)$} and \textit{$F_{M}^l\left( {{x_k}} \right)$} as follow:
	\begin{equation}
		L\left( {{x_k}} \right) = \sum\limits_{l = 1}^L {\frac{1}{{{w_l} \cdot {h_l}}}\sum\limits_{i = 1}^{{w_l}} {\sum\limits_{j = 1}^{{h_l}} {p_{\left( {i,j} \right)}^l\left( {{x_k}} \right)} } },
	\end{equation}
where \textit{L} represents total number of critical layers.

\subsection{Computation of the anomaly map}\label{subsec34}

In training phase, \textit{Trans} has stronger fitting ability in anomaly-free images than the AGN under the guidance of AGN. However, AGN is knowledgeable on anomaly images because of the strong generalization\footnote{However, AGN is knowledgeable on anomaly images because of the strong generalization, while Trans is unfamiliar with such images. This due to that AGN is pre-trained on ImageNet, which can generalize well across datasets \cite{kornblith2019better}, while \textit{Trans} is trained from scratch.}, while \textit{Trans} is unfamiliar with such images. Therefore, the abnormal regions of anomaly images show a large deviation from the training data manifold of \textit{Trans} when the anomaly images are input into AGN and \textit{Trans}, respectively. Given a testing dataset \textit{${D_{test}}{\rm{ = }}\left\{ {{y_1},{y_2},...,{y_M}} \right\}$} consisting both anomaly and normal images, we assign an anomaly map \textit{$M\left( {{y_k}} \right) \in {\mathbb{R}^{H \times W}}$} to a test image \textit{${y_k} \in {D_{test}}$} for \textit{$k \in \left[ {1,M} \right]$}, where \textit{H} and \textit{W} are the height and width of the test image. The anomaly score of pixel at position \textit{$\left( {i,j} \right)$} of $i \in \left[ {1,{H}} \right]$, $j \in \left[ {1,{W}} \right]$ indicates the deviation degree of pixel at position  \textit{$\left( {i,j} \right)$} from the data manifold of anomaly-free images. We utilize the anomaly map \textit{$M\left( {{y_k}} \right)$} to realize anomaly detection and localization. Next, we show the specific computation of anomaly map.

    We input a test image \textit{${y_k}$} into our model to obtain the group of feature maps \textit{${F_G}({y_k}){\rm{ = }}\left\{ {F_G^1({y_k}),F_G^2({y_k}),...,F_G^L({y_k})} \right\}$} and \textit{${F_{M}}({y_k}){\rm{ = }}\left\{ {F_{M}^1({y_k}),F_{M}^2({y_k}),...,F_{M}^L({y_k})} \right\}$} extracted from different critical layers of AGN and \textit{Trans}, respectively.

We get a group of loss \textit{$P{\rm{ = }}\left\{ {{P^1},{P^2},...,{P^L}} \right\}$} by computing pixel-wise \textit{${L_2}$} loss between the feature map from the same layers at \textit{${F_G}({y_k})$} and \textit{${F_{M}}({y_k})$} as follows:

	\begin{equation}
		{P^l} = \frac{1}{{{w_l} \cdot {h_l}}}\sum\limits_{i = 1}^{{c_l}} {\frac{1}{2}\left\| {F_G^l\left( {{y_k}} \right) - F_{M}^l\left( {{y_k}} \right)} \right\|_2^2}  \in {\mathbb{R}^{{w_l} \times {h_l}}},
	\end{equation}
where \textit{${P^l} \in P$} denotes the pixel-wise loss of \textit{l}-th critical layer between AGN and \textit{Trans}. We assign a weight to different critical layers because the different critical layers correspond to anomalies at various sizes, respectively. Here, we utilize MSE loss and cosine similarity metric of feature map in \textit{l}-th critical layer to define two coefficients \textit{$\alpha _{mse}^l$} and \textit{$\alpha _{{\rm{cos}}}^l$} as follows:
	\begin{equation}
		\alpha _{mse}^l = \frac{1}{{{w_l} \cdot {h_l}}}{\sum\limits_{i = 1}^{{w_l}} {\sum\limits_{j = 1}^{{h_l}} {\left( {F_G^l{{\left( {{y_k}} \right)}_{(i,j)}} - F_{M}^l{{\left( {{y_k}} \right)}_{\left( {i,j} \right)}}} \right)} } ^2}
	\end{equation}
and

    \begin{equation}
    		\alpha _{{\rm{cos}}}^l = 1 - \frac{{vec{{\left( {F_G^l\left( {{y_k}} \right)} \right)}^T} \cdot vec\left( {F_{M}^l\left( {{y_k}} \right)} \right)}}{{\left\| {vec\left( {F_G^l\left( {{y_k}} \right)} \right)} \right\|\left\| {vec\left( {F_{M}^l\left( {{y_k}} \right)} \right)} \right\|}}.
    	\end{equation}

    Here, \textit{$vec\left(  \cdot  \right)$} is a vectorization function transforming a matrix with arbitrary dimensions into a 1-D vector.
	
\textit{$\alpha _{mse}^l$} and \textit{$\alpha _{{\rm{cos}}}^l$} denote the similarity in spatial distance and direction of \textit{l}-th feature maps from AGN and \emph{Trans}. The larger the values of \textit{$\alpha _{mse}^l$} and \textit{$\alpha _{{\rm{cos}}}^l$} are, the less similarity the \emph{l}-th feature maps are in spatial distance and direction, and the greater the possibility of anomalies. We effectively combine the similarity of feature maps in spatial distance and direction, and use harmonic mean of \textit{$\alpha _{mse}^l$} and \textit{$\alpha _{{\rm{cos}}}^l$} as the weight of \emph{l}-th feature maps. \textit{${\alpha ^l}$} is formulated as
	\begin{equation}
		{\alpha ^l} = \frac{{\lambda  \cdot \alpha _{{\rm{cos}}}^l \cdot \alpha _{mse}^l}}{{\alpha _{mse}^l + \lambda  \cdot \alpha _{{\rm{cos}}}^l}},
	\end{equation}
where \textit{$\lambda $} is set to make the scale of \textit{$\alpha _{mse}^l$} and \textit{$\alpha _{{\rm{cos}}}^l$} the same. Our goal is to assign high weight to feature maps with low similarity, so as to get high anomaly score with anomalous regions eventually. Therefore, the anomaly map is achieved by
	\begin{equation}
		M\left( {{y_k}} \right) = {G_\sigma }\left( {\sum\limits_{l = 1}^L {{\alpha ^l}R\left( {{P^l}} \right)} } \right) \in {\mathbb{R}^{H \times W}}.
	\end{equation}

	Here, \textit{$R\left(  \cdot  \right)$} resizes the elements of \textit{$P{\rm{ = }}\left\{ {{P^1},{P^2},...,{P^L}} \right\}$} to the spatial size of \textit{$\left( {H \times W} \right)$}. \textit{${G_\sigma }\left(  \cdot  \right)$} represents a Gaussian filter with standard deviation of \textit{$\sigma $}. Finally, we define the region with high anomaly score in \textit{$M\left( {{y_k}} \right)$} as anomalous regions and the maximum value of \textit{$M\left( {{y_k}} \right)$} as the final anomaly score of the test image \textit{${y_k}$} for anomaly detection.

\section{Experiments}\label{sec4}

 In this section, we first elaborate on the details of \textit{GTrans} structure and its parameters. Second, to demonstrate the effectiveness of our approach, we compare our method to state-of-the-art results on the MVTec AD benchmark, considering in image-level and pixel-level, respectively.

\subsection{Datesets and metrics}\label{subsec41}

\subsubsection{Datesets}\label{subsubsec411}

We evaluate \textit{GTrans} on the MVTec AD dataset\footnote{The datasets analysed during the current study are available at https://www.mvtec.com/company/research/datasets/mvtec-ad.} which contains over 5000 high-resolution images divided into fifteen different object and texture categories. Each category of the MVTec AD dataset comprises a set of anomaly-free training images and a test set of images with various kinds of anomalies as well as images without anomaly.

\subsubsection{Metrics}\label{subsubsec412}

We use the Area Under the Receiver Operating Characteristic curve (AUROC) and Area Under the Per-Region-Overlap Curve (AUPRO) to evaluate the performence of \textit{GTrans}. AUROC is computed on different levels of threshold in favor of large anomalous regions. AUPRO computes a threshold-independent evaluation metric based on the Per-Region-Overlap (PRO) to better account for large and small anomalies in localization. Only the standard AUROC provides image-level anomaly detection.

\subsection{Implementation Details}\label{subsec42}

We resize all images in the MVTec AD dataset to \textit{$256 \times 256$} and center crop them to \textit{$224 \times 224$}. We take the preprocessed images of MVTec as the input of \textit{GTrans}. We divided eighty percent of the MVTec AD dataset into the training set and twenty percent into the validation set without additional data enhancement. In the training phase, \textit{GTrans} is trained only on the normal images with a batch size of 32 for 300 epochs. We utilize Adam optimizer with initial learning rate \textit{${10^{ - 3}}$} and weight decay \textit{${10^{ - 4}}$} for optimization. To enhance the stability of the \textit{GTrans}, we define a large initial learning rate to improve the convergence speed of \textit{GTrans}, and then the learning rate decreases with the increase of the number of iterations gradually. We utilize exponential decay equation to realize the decay of learning rate as
	\begin{equation}
		lr = l{r_{init}} \times rat{e^{\frac{{step}}{{totalstep}}}},
	\end{equation}
where \textit{$l{r_{init}}$} and \textit{$lr$} represent the initial and current learning rate respectively, \textit{$rate$} is decay factor with a  value of 0.9, \textit{$step$} and \textit{$totalstep$} denote the current and total iteration number respectively. We implement our method in PyTorch and conduct all experiments on a machine equipped with an Intel i9-9900X and an NVIDIA GeForce RTX 2080 Ti GPU.

\subsection{Architecture Details}\label{subsec43}

As shown in Figure \ref{FIG:2}, \textit{GTrans} consists of AGN and \emph{Trans}. We illustrate the parameter selection of \emph{GTrans} below.

\subsubsection{AGN}\label{subsubsec431}

We extract features with a ResNet-34 (R34) and a Wide ResNet-50-2 (WR50). All backbones of AGN pre-trained on ImageNet, and the parameters of AGN are frozen during the training phase. As to the position of the selected guide feature, we choose features extracted by three intermediate layer groups, that is \textit{$l = 3$}.

\subsubsection{\emph{Trans}}\label{subsubsec432}

\emph{Trans} is composed of a backbone network, a Tokenizer, a TFM and a mapper. We keep the backbone of \emph{Trans} consistent with AGN. For Tokenizer, inspired by \cite{wu2020visual}, we use a few visual tokens to generalise semantic concepts. Hence, we set the number of semantic groups \textit{$g = 8$} and dimension \textit{$d = 256$}. For TFM, we utilize as few as two decoders and encoders to achieve superior performance. After mapper, the \emph{Trans} and AGN output three feature maps with the size \textit{$56 \times 56$}, \textit{$28 \times 28$} and \textit{$14 \times 14$}, respectively.

\subsection{Results}\label{subsec44}

In this section, we show the result of our method. To demonstrate the feasibility and effectivity of our experiment, we conduct anomaly detection and localization using the MVTec AD dataset with anomalies. We present the AUROC and AUPRO score for each category to give an intuition of the effect of \emph{GTrans} on different categories. We compare our method to state-of-the-art results on the MVTec AD benchmark.

\begin{table*}[!t]
		\newcommand{\tabincell}[2]{\begin{tabular}{@{}#1@{}}#2\end{tabular}}
		\begin{center}\caption{The AUROC results for anomaly detection on the MVTec AD dataset using AUROC\%.}
			\centering\label{table1}
            \resizebox{\linewidth}{!}{
			\begin{tabular}{lcccccccccc}
				\hline
				\multicolumn{1}{l|}{Category}                                         & AnoGAN\cite{schlegl2017unsupervised} & AE-SSIM\cite{bergmann2018improving} & GANomaly\cite{akcay2018ganomaly} & SPADE\cite{cohen2020sub} & Patch-SVDD\cite{yi2020patch} & CutPaste\cite{li2021cutpaste}       & PaDiM\cite{defard2021padim} & InTra\cite{pirnay2021inpainting}          & \emph{GTrans}-R34      & \emph{GTrans}-WR50      \\ \hline
				\multicolumn{1}{l|}{carpet}                      & 49     & 67      & 69.9     & -     & 92.9       & 93.1           & -     & 98.8           & 99.9           & \textbf{100.0} \\
				\multicolumn{1}{l|}{grid}                        & 51     & 69      & 70.8     & -     & 94.6       & 99.9           & -     & \textbf{100.0} & 99.4           & 99.0           \\
				\multicolumn{1}{l|}{leather}                     & 52     & 46      & 84.2     & -     & 90.9       & \textbf{100.0} & -     & \textbf{100.0} & \textbf{100.0} & \textbf{100.0} \\
				\multicolumn{1}{l|}{tiletable2}               & 51     & 52      & 79.4     & -     & 97.8       & 93.4           & -     & 98.2           & \textbf{98.9}  & \textbf{98.9}  \\
				\multicolumn{1}{l|}{wood}                        & 68     & 83      & 83.4     & -     & 96.5       & 98.6           & -     & 98.0           & \textbf{99.9}  & 99.5           \\ \hline
				\multicolumn{1}{l|}{\textbf{avg.textures}}       & 54.2   & 63.4    & 77.5     & -     & 94.5       & 97.0           & 99.0  & 99.0           & \textbf{99.6}  & 99.5           \\ \hline
				\multicolumn{1}{l|}{bottle}                      & 69     & 88      & 89.2     & -     & 98.6       & 98.3           & -     & \textbf{100.0} & \textbf{100.0} & \textbf{100.0} \\
				\multicolumn{1}{l|}{cable}                       & 53     & 61      & 75.7     & -     & 90.3       & 80.6           & -     & 84.2           & \textbf{99.9}  & 99.7           \\
				\multicolumn{1}{l|}{capsule}                     & 58     & 61      & 73.2     & -     & 76.7       & 96.2           & -     & 86.5           & \textbf{97.0}  & 95.1           \\
				\multicolumn{1}{l|}{hazelnut}                    & 50     & 54      & 78.5     & -     & 92.0       & 97.3           & -     & 95.7           & \textbf{100.0} & \textbf{100.0} \\
				\multicolumn{1}{l|}{metal nut}                   & 50     & 54      & 70.0     & -     & 94.0       & 99.3           & -     & 96.9           & 99.5           & \textbf{100.0} \\
				\multicolumn{1}{l|}{pill}                        & 62     & 60      & 74.3     & -     & 86.1       & 92.4           & -     & 90.2           & \textbf{93.9}  & 92.7           \\
				\multicolumn{1}{l|}{screw}                       & 35     & 51      & 74.6     & -     & 81.3       & 86.3           & -     & 95.7           & \textbf{98.3}  & 93.6           \\
				\multicolumn{1}{l|}{toothbrush}                  & 57     & 74      & 65.3     & -     & 100.0      & 98.3           & -     & 99.7           & \textbf{100.0} & \textbf{100.0} \\
				\multicolumn{1}{l|}{transistor}                  & 67     & 52      & 79.2     & -     & 91.5       & 95.5           & -     & 95.8           & \textbf{99.7}  & 99.3           \\
				\multicolumn{1}{l|}{zipper}                      & 59     & 80      & 74.5     & -     & 97.9       & 99.4           & -     & 99.4           & 98.9           & \textbf{99.9}  \\ \hline
				\multicolumn{1}{l|}{\textbf{avg.objects}}        & 56.0   & 63.5    & 75.5     & -     & 90.8       & 94.4           & 97.2  & 94.4           & \textbf{98.7}  & 98.0           \\ \hline
				\multicolumn{1}{l|}{\textbf{avg.all categories}} & 55.4   & 63.5    & 76.1     & 85.5  & 92.1       & 95.2           & 97.9  & 95.9           & \textbf{99.0}  & 98.5           \\ \hline
			\end{tabular} }
		\end{center}
	\end{table*}	

\begin{table*}[!t]
		\newcommand{\tabincell}[2]{\begin{tabular}{@{}#1@{}}#2\end{tabular}}
		\begin{center}\caption{Comparison of our models with the state-of-the-art for the anomaly localization on the MVTec AD dataset. Results are displayed as tuples(AUROC\%, PRO-SCORE\%).}
			\centering\label{table2}
            \resizebox{\linewidth}{!}{
			\begin{tabular}{lcccccccccc}
				\hline
				\multicolumn{1}{l|}{Category}                    & AnoGAN\cite{schlegl2017unsupervised}       & AE-SSIM\cite{bergmann2018improving}    & Patch-SVDD\cite{yi2020patch} & Student\cite{bergmann2020uninformed}   & CutPaste\cite{li2021cutpaste}  & SPADE\cite{cohen2020sub}        & InTra\cite{pirnay2021inpainting}             & \emph{GTrans}-R34    & \emph{GTrans}-WR50    \\ \hline
				\multicolumn{1}{l|}{carpet}                      & (54, 20.4)   & (87, 64.7) & (92.6, -)  & (-, 69.5) & (98.3, -) & (97.5, 94.7) & (98.8, 95.8)      & (\textbf{99.2}, \textbf{97.3}) & (\textbf{99.2}, 96.7) \\
				\multicolumn{1}{l|}{grid}                        & (58, 22.6)   & (94, 84.9) & (96.2, -)  & (-, 81.9) & (97.5, -) & (93.7, 86.7) & (\textbf{99.0}, \textbf{96.6})      & (98.8, 93.3) & (\textbf{99.0}, 96.2) \\
				\multicolumn{1}{l|}{leather}                     & (64, 37.8)   & (78, 56.1) & (97.4, -)  & (-, 81.9) & (99.5, -) & (97.6, 97.2) & (99.3, 98.0)      & (99.4, 97.8) & (\textbf{99.5}, \textbf{98.4}) \\
				\multicolumn{1}{l|}{tile}                        & (50, 17.7)   & (59, 17.5) & (91.4, -)  & (-, 91.2) & (90.5, -) & (87.4, 75.9) & (\textbf{97.4}, \textbf{92.1})      & (95.0, 79.2) & (97.0, 84.5) \\
				\multicolumn{1}{l|}{wood}                        & (62, 38.6)   & (73, 60.5) & (90.8, -)  & (-, 72.5) & (95.5, -) & (88.5, 87.4) & (97.2, 93.6)      & (97.1, 92.2) & (\textbf{97.8}, \textbf{95.7}) \\ \hline
				\multicolumn{1}{l|}{\textbf{avg.textures}}       & (57.6, 27.4) & (78, 56.7) & (93.7, -)  & (-, 79.4) & (96.3, -) & (92.9, 88.4) & (98.3, \textbf{95.2})      & (97,9, 92.0) & (\textbf{98.5}, 94.3) \\ \hline
				\multicolumn{1}{l|}{bottle}                      & (86, 62.0)   & (93, 83.4) & (98.1, -)  & (-, 91.8) & (97.6, -) & (98.4, \textbf{95.5}) & (\textbf{98.8}, 95.1)      & (97.9, 92.6) & (98.5, 94.9) \\
				\multicolumn{1}{l|}{cable}                       & (78, 38.3)   & (82, 47.8) & (96.8, -)  & (-, 86.5) & (90.0, -) & (97.2, 90.9) & (95.5, 87.7)      & (97.3, 91.4) & (\textbf{97.6}, \textbf{92.3}) \\
				\multicolumn{1}{l|}{capsule}                     & (84, 30.6)   & (94, 86.0) & (95.8, -)  & (-, 91.6) & (97.4, -) & (\textbf{99.0}, \textbf{93.7}) & (98.3, 92.2)      & (98.1, 77.0) & (98.0, 86.9) \\
				\multicolumn{1}{l|}{hazelnut}                    & (87, 69.8)   & (97, 91.6) & (97.5, -)  & (-, 93.7) & (97.3, -) & (\textbf{99.1}, 95.4) & (98.5, 94.3)      & (98.8, 98.1) & (98.8, \textbf{98.7}) \\
				\multicolumn{1}{l|}{metal nut}                   & (76, 32.0)   & (89, 60.3) & (98.0, -)  & (-, 89.5) & (93.1, -) & (\textbf{98.1}, 94.4) & (97.6, \textbf{94.5})      & (97.7, 93.0) & (\textbf{98.1}, 94.3) \\
				\multicolumn{1}{l|}{pill}                        & (87, 77.6)   & (91, 83.0) & (95.1, -)  & (-, 93.5) & (95.7, -) & (96.5, 94.6) & (97.8, \textbf{96.5})      & (98.6, 94.2) & (\textbf{98.9}, 95.1) \\
				\multicolumn{1}{l|}{screw}                       & (80, 46.6)   & (96, 88.7) & (95.7, -)  & (-, 92.8) & (96.7, -) & (98.9, \textbf{96.0}) & (98.3, 93.0)      & (\textbf{99.2}, 93.2) & (\textbf{99.2}, 95.5) \\
				\multicolumn{1}{l|}{toothbrush}                  & (93, 74.9)   & (92, 78.4) & (98.1, -)  & (-, 86.3) & (98.1, -) & (97.9, \textbf{93.5}) & (\textbf{98.9}, 92.2)      & (98.3, 87.6) & (98.6, 89.7) \\
				\multicolumn{1}{l|}{transistor}                  & (86, 54.9)   & (90, 72.5) & (\textbf{97.0}, -)  & (-, 70.1) & (93.0, -) & (94.1, \textbf{87.4}) & (82.5, 69.5)      & (95.8, 82.3) & (94.1, 79.9) \\
				\multicolumn{1}{l|}{zipper}                      & (78, 46.7)   & (88, 66.5) & (95.1, -)  & (-, 93.3) & (\textbf{99.3}, -) & (96.5, 92.6) & (98.5, 95.2)      & (98.0, 94.0) & (98.8, \textbf{95.6}) \\ \hline
				\multicolumn{1}{l|}{\textbf{avg.objects}}        & (83.5, 53.3) & (91, 75.8) & (96.7, -)  & (-, 88.9) & (95.8, -) & (97.6, \textbf{93.4}) & (96.5, 91.0)      & (98.0, 90.3) & (\textbf{98.1}, 92.3) \\ \hline
				\multicolumn{1}{l|}{\textbf{avg.all categories}} & (74.9, 44.7) & (87, 69.4) & (95.7, -)  & (-, 85.7) & (96.0, -) & (96.5, 91.7) & (97.0, 92.1) & (97.9, 90.9) & (\textbf{98.2}, \textbf{93.0}) \\ \hline
			\end{tabular} }
		\end{center}
	\end{table*}

\subsubsection{Detection}\label{subsubsec441}

Table \ref{table1} presents the AUROC results for anomaly detection on the MVTec AD dataset. We take the maximum value of anomaly map issued by \textit{GTrans} (see Section \ref{subsec34}) for the anomaly detection and report standard AUROC as a detection metric. Since the other baselines have different backbones, we try a R34 and a WR50 as the backbone of our model respectively. As shown in Table \ref{table1}, \textit{GTrans}-R34 outperforms the other methods by 1.1\% to 43.6\% in AUROC score on average for all the categories.

\subsubsection{Localization}\label{subsubsec442}

Table \ref{table2} shows the AUROC and the PRO-score results for anomaly localization on the MVTec AD dataset. As shown in Table \ref{table2}, \textit{GTrans}-WR50 outperforms all the other methods in both the AUROC and the PRO-score on average for all the categories. Compared with the method based on Transformer, the performance of our method is 1.2\% higher than InTra\cite{pirnay2021inpainting}. For PRO-score, InTra\cite{pirnay2021inpainting} and SPADE\cite{cohen2020sub} achieve the best performance in the texture and object categories respectively. However, our method performs well in both texture and object categories, outperforming InTra\cite{pirnay2021inpainting} by 0.9\% and SPADE\cite{cohen2020sub} 1.3\% in PRO-score on average for all the categories. When we further analyze the performance of \textit{GTrans}-WR50, we find that \textit{GTrans}-WR50 outperforms Student\cite{bergmann2020uninformed} by 7.3\% in PRO-score, indicating the effectiveness of our method in multi-scale feature fusion and information interaction.

\section{Ablation Studies}\label{sec5}

\subsection{Intermediate Knowledge}\label{subsec51}

\begin{table}[!t]
		\newcommand{\tabincell}[2]{\begin{tabular}{@{}#1@{}}#2\end{tabular}}
		\begin{center}\caption{Study of the anomaly detection and localization performance with a R34 backbone using different feature layers. results are displayed as tuples(detection AUROC\%, location AUROC\%) on MVTec AD dataset.}
			\centering\label{table3}
			\begin{tabular}{lccc}
                \hline
				\multicolumn{1}{l|}{Layer used}    & avg.textures & avg.objects     & avg.all categories   \\ \hline
				\multicolumn{1}{l|}{Layer 1+2+3+4} & (98.4, 91.5)        & (97.3, 93.2)          & (97.6, 92.8)   \\
				\multicolumn{1}{l|}{Layer 1+2+3}   & (\textbf{99.6}, \textbf{96.0})        & (\textbf{97.9}, \textbf{96.}7)          & (\textbf{98.3}, \textbf{96.5})   \\
				\multicolumn{1}{l|}{Layer 2+3+4}   & (98.9, 92.2)        & (97.1, 94.2)          & (97.6, 93.7)   \\
				\multicolumn{1}{l|}{Layer 2+3}     & (\textbf{99.6}, 95.6)        & \textbf{(\textbf{97.9}, 96.5)} & (\textbf{98.3}, 96.2)   \\ \hline
			\end{tabular}
		\end{center}
	\end{table}

In this part, we evaluate the impact of the position of selected guide features in \textit{GTrans}. In Table \ref{table3}, we show the performance of anomaly detection and localization on the MVTec AD dataset of \textit{GTrans} with a R34 backbone when using different combination critical layers (Layer 1+2+3+4, Layer 1+2+3, Layer 2+3+4, Layer 3+4). The final anomalous maps of all experiments in this part simply add the anomalous maps of different layers. It is known that feature layers of CNN can express various levels of abstract information. Bottom layers tend to extract low-level information such as textures, while top layers pay attention to low-resolution features that contain semantic information. It can be observed from Table \ref{table3} that Layer 1+2+3 can achieve the best performance in both texture and object categories. From the comparison of Layer 1+2+3+4 and Layer 1+2+3, it can be seen that the performance of anomaly detection and localization declines when the anomalous map generated by the fourth critical layer is added. The reason is that the features of the top layers are low-resolution, leading to the rough segmentation in detail texture. According to the experimental results of Layer 2+3, the information of the intermediate layers can express and extract anomalies to a large extent. When the information of the shallow layer is added on this basis, the model can extract the anomalies and pay attention to the processing of edge texture, improving the performance of the model.

\subsection{The structure of TFM}\label{subsec52}

This part evaluates the impact of an added decoder and the number of TFM blocks on experimental results.

    In Table \ref{table_tfm}, we show the anomaly detection and localization performance on the MVTec AD dataset of \textit{GTrans} with an R34 backbone. It can be seen that the detection AUROC of the model with TFM  is 2.7\% higher than that without TFM, while for localization AUROC, it can achieve 1.9\% higher than the latter model.

In Table \ref{table4}, we show the performance in anomaly detection and localization on the MVTec AD dataset of \textit{GTrans} with a R34 backbone when using the different structure of TFM in \emph{Trans}, where \textit{S} represents the number of TFM blocks. For structure which added decoder, we decode the output of the encoder through the features extracted from the AGN.

    From Table \ref{table4}, we can notice that for the same number of TFM blocks, the structure which added decoder outperforms the structure of pure encoder on MVTec AD dataset by 0.2\% to 0.4\% in the detection AUROC and 0.2\% in localization AUROC. On this basis, we conduct experiments with the different number of TFM blocks for the two structures. The experimental results indicate that the number of TFM blocks has little effect on the structure of the pure encoder. For a structure that added decoder, the performance in detection of the structure with \textit{S}=2 is 0.3\% higher than the structure with \textit{S}=1, while the performance in localization is basically in a stable state. When \textit{S}\textgreater 2, the performance in detection and localization remain stable or slightly decrease, indicating that the structure with \textit{S}=2 is enough to fit the distribution manifold of normal data and detect anomalies.

    \begin{table}[!t]
    		\newcommand{\tabincell}[2]{\begin{tabular}{@{}#1@{}}#2\end{tabular}}
    		\begin{center}\caption{Performance with a R34 backbone with/without TFM architecture. Results are displayed as tuples(detection AUROC\%, location AUROC\%) on MVTec AD dataset.}
    			\centering\label{table_tfm}
    			\begin{tabular}{cccc}
    				\hline
    				\multicolumn{1}{c|}{Structure}    & avg.textures & avg.objects & avg.all categories  \\ \hline
    				\multicolumn{1}{c|}{without TFM}             & (99.4, 96.8)        & (93.2, 97.3)       & (95.2, 97.1) \\
    				\multicolumn{1}{c|}{with TFM}        & (\textbf{99.6}, \textbf{97.9})        & (\textbf{98.7}, \textbf{98.0})       & (\textbf{97.9}, \textbf{99.0}) \\
    			 \hline
    			\end{tabular}
    		\end{center}
    	\end{table}

	\begin{table}[!t]
		\newcommand{\tabincell}[2]{\begin{tabular}{@{}#1@{}}#2\end{tabular}}
		\begin{center}\caption{Study of performance with a R34 backbone using different structure of TFM. results are displayed as tuples(detection AUROC\%, location AUROC\%) on MVTec AD dataset.}
			\centering\label{table4}
			\begin{tabular}{lcccc}
                \hline
				\multicolumn{1}{l|}{Structure}             & block & avg.textures   & avg.objects & avg.all categories   \\ \hline
				\multicolumn{1}{c|}{\multirow{3}{*}{pure encoder}} & S=1   & (99.4, 97.9)          & (98.1, 97.6)       & (98.5, 97.7) \\
				\multicolumn{1}{c|}{}                            & S=2   & (99.3, 97.9)          & (98.2, 97.7)       & (98.6, 97.7) \\
				\multicolumn{1}{c|}{}                            & S=3   & (99.3, 97.9)          & (98.2, 97.6)       & (98.6, 97.7)  \\ \hline
				\multicolumn{1}{c|}{\multirow{3}{*}{added decoder}}    & S=1   & (99.4, 97.9) & (98.3, 97.9)       & (98.7, \textbf{97.9})  \\
				\multicolumn{1}{c|}{}                            & S=2   & (\textbf{99.6}, 97.9)          & (\textbf{98.7}, \textbf{98.0})       & (\textbf{99.0}, \textbf{97.9})   \\
				\multicolumn{1}{c|}{}                            & S=3   & (99.4, \textbf{98.0})          & (\textbf{98.7}, 97.9)       & (98.9, \textbf{97.9})  \\ \hline
			\end{tabular}
		\end{center}
	\end{table}

\subsection{Computation of anomaly map}\label{subsec53}

	This part evaluates the influence of different function of anomaly score and fusion mode of anomaly map on experimental results. Table \ref{table5} shows the performance in detection and localization with a R34 backbone using different function of anomaly map, where \textit{$\alpha$} represents the weight of the anomaly map of our model, \textit{$\alpha _{mse}$} and \textit{$\alpha _{cos}$} denote the difference value in spatial distance and direction of anomaly map (see section \ref{subsec34}). As can be seen from Table \ref{table5}, the weight coefficient obtained according to the importance of the anomaly map outperforms the fixed coefficient by 0.9\% in detection AUROC and 0.1\% in localization AUROC. \textit{$\alpha _{mse}$} pays more attention to distance differences and performs better in detecting subtle anomalies, while \textit{$\alpha _{cos}$} focuses on direction differences and performs better in detecting diversity anomalies. Therefore, we use the harmonic mean values of \textit{$\alpha _{mse}$} and \textit{$\alpha _{cos}$} as the coefficient of the final anomaly score to detect anomalies with various sizes better.
	
\begin{figure}[!t]
		\centering
		\includegraphics[width=11cm]{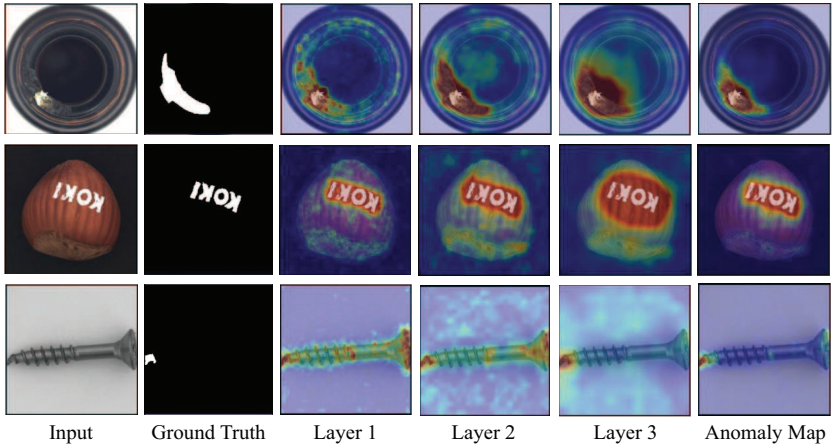}
		\caption{Visualization of anomaly samples from bottle, hazelnut and screw of MVTec AD dataset. Columns from left to right correspond to anomaly sample, ground truth, anomaly maps generated by three layers (Layer 1, Layer 2, Layer 3), and the final anomaly maps respectively.}
		\label{FIG:4}
	\end{figure}

	As can be seen from the Figure \ref{FIG:4}, anomaly maps extracted from different semantic layers express different abstract information. The anomaly map extracted from the shallow layer (Layer 1) pays more attention to texture and edge extraction but has weak semantic information. The anomaly map extracted from the middle layer (Layer 2) is more carefully segmented than shallow layer but easily affected by background noise. While the anomaly map extracted from deep layer (Layer 3) focuses on semantic information but has low resolution, resulting in poor segmentation. We attempt to combine the anomaly maps extracted from different semantic layers to obtain accurate pixel-level localization of anomalies.

Table \ref{table6} shows the performance in detection and localization with a R34 backbone using the different combinations of anomaly maps, where P1 to P5 represent different combination modes, the Arabic numerals represent anomalous maps generated by different critical layers. It can be observed that the performance of anomaly map fusion is 0.4\% to 1.1\% and 0.8\% to 2.3\% higher than that of single-layer anomaly map in detection and localization AUROC respectively, indicating that the effective fusion of anomaly map with different levels is conducive to detecting and locating anomalies with different sizes.

\begin{table}[!t]
		\newcommand{\tabincell}[2]{\begin{tabular}{@{}#1@{}}#2\end{tabular}}
		\begin{center}\caption{Study of performance with a R34 backbone using different function of anomaly maps. results are displayed as tuples(detection AUROC\%, location AUROC\%) on MVTec AD dataset.}
			\centering\label{table5}
			\begin{tabular}{cccc}
				\hline
				\multicolumn{1}{c|}{Coefficient}    & avg.textures & avg.objects & avg.all categories  \\ \hline
				\multicolumn{1}{c|}{0.5}             & (98.7, \textbf{98.0})        & (97.6, 97.8)       & (98.0, 97.8) \\
				\multicolumn{1}{c|}{\textit{$\alpha _{mse}$}}        & (99.5, 97.7)        & (\textbf{98.7}, \textbf{98.0})       & (98.9, \textbf{97.9}) \\
				\multicolumn{1}{c|}{\textit{$\alpha _{cos}$}}     & (\textbf{99.6}, 97.8)        & (98.6, 97.9)       & (98.9, \textbf{97.9}) \\
				\multicolumn{1}{c|}{\textit{$\alpha$}} & (\textbf{99.6}, 97.9)        & (\textbf{98.7}, \textbf{98.0})       & (\textbf{99.0}, \textbf{97.9}) \\ \hline
			\end{tabular}
		\end{center}
	\end{table}

	\begin{table}[!t]
		\newcommand{\tabincell}[2]{\begin{tabular}{@{}#1@{}}#2\end{tabular}}
		\begin{center}\caption{Study of performance with a R34 backbone using different combination of anomaly maps. results are displayed as tuples(detection AUROC\%, location AUROC\%) on MVTec AD dataset.}
			\centering\label{table6}
			\begin{tabular}{cccc}
				\hline
				\multicolumn{1}{l|}{Combination Mode}               & avg.textures  & avg.objects & avg.all categories  \\ \hline
				\multicolumn{1}{l|}{P1: 3}              & (99.3, 95.2)        & (97.2, 96.0)       & (97.9, 95.7) \\
				\multicolumn{1}{l|}{P2: 1*3}            & (99.5, \textbf{98.2})        & (98.0, 97.9)       & (98.5, \textbf{98.0}) \\
				\multicolumn{1}{l|}{P3: 2*3}            & (99.5, 97.3)        & (98.4, 97.8)       & (98.8, 97.6) \\
				\multicolumn{1}{l|}{P4: 1+2+3} & (99.5, 95.6)        & (97.8, 96.9)       & (98.3,96.5)  \\
				\multicolumn{1}{l|}{P5: 1*2*3}          & (99.5, \textbf{98.2})        & (98.4, 97.8)       & (98.8, 97.9) \\
				\multicolumn{1}{l|}{P6: 1*3+2*3}        & (\textbf{99.6}, 97.9)        & (\textbf{98.7}, \textbf{98.0})       & (\textbf{99.0}, 97.9) \\ \hline
			\end{tabular}
		\end{center}
	\end{table}

As can be noticed from Table \ref{table6} that the performance in detection of P3 is 0.3\% higher than that of P2, and the performance in localization is 0.4\% lower than that of P2. Qualitatively, different levels of anomaly maps detect anomalies with different sizes. P2 mode performs well with anomalies of texture categories because providing more edge information, while P3 mode works well with object categories because providing more semantic information. To balance the characteristics of different anomalies, we effectively combined P2 and P3 into P6, resulting in optimal performance.

\section{Conclusions}\label{sec6}

    We propose a multiresolution feature guidance method based on Transformer named \emph{GTrans} for unsupervised anomaly detection and localization. First,  \emph{GTrans} utilizes the difference in generalization ability between AGN and \emph{Trans} to locate anomalous regions.
    Second, we utilize the TFM module modified by Transformer to enhance the information interaction ability of multi-scale features.
    Also, for anomaly maps, we propose a new generation function that jointly considers spatial and directional distances.
    Finally, we conduct a series of ablation studies to demonstrate the effectiveness of \emph{GTrans}. Experimental results on MVTec AD dataset show that \textit{GTrans} can achieve the state-of-the-art performance in both detection and localization.

\bibliography{sn-bibliography}


\end{document}